# Target Recognition Algorithm for Monitoring Images in Electric Power Construction Process

Hao Song, Wei Lin, Wei Song, Man Wang

School of Information Science and Engineering, Chongqing Jiaotong University

**Abstract**

To enhance precision and comprehensiveness in identifying targets in electric power construction monitoring video, a novel target recognition algorithm utilizing infrared imaging is explored. This algorithm employs a color processing technique based on a local linear mapping method to effectively recolor monitoring images. The process involves three key steps: color space conversion, color transfer, and pseudo-color encoding. It is designed to accentuate targets in the infrared imaging. For the refined identification of these targets, the algorithm leverages a support vector machine approach, utilizing an optimal hyperplane to accurately predict target types. We demonstrate the efficacy of the algorithm, which achieves high target recognition accuracy in both outdoor and indoor electric power construction monitoring scenarios. It maintains a false recognition rate below 3% across various environments.

## 1  Introduction

The imaging mechanism of infrared imaging is relatively unique, and it has been used in a large number of military, national defense, electric power, and hospitals in the field of surveillance work. In electric power construction work, electric power construction video surveillance image target recognition is very critical to the safety of electric power construction personnel and equipment management has a certain significance [1-2]. At present, there are a large number of research results for the image target recognition problem, for example, the literature [3] will use the non-negative matrix decomposition method in spatial target image recognition to propose the spatial image target recognition algorithm based on the non-negative matrix decomposition, although the algorithm can identify the target of the video surveillance image of the electric power construction, the recognition accuracy of the algorithm is not high because the algorithm can not implement the colorization of the video surveillance image of the electric power construction, the algorithm can not be used in the video surveillance image of the electric power construction, the algorithm is based on infrared imaging.

In the infrared image obtained from the power construction video surveillance based on infrared imaging, the significant feature of the target is the target temperature field, which is constrained by the performance of the infrared detector and the related technology The monitoring image of the infrared thermal image application monitoring system is a black-and-white image The significance of the target in the black-and-white image is low, so that the algorithm does not have a high accuracy of the recognition of the target in the image; Literature [4] proposes the image detection algorithm of the significant target based on the depth learning, which can realize the power construction video surveillance image, and it can realize the power construction video surveillance image of the significant target. Literature [4] proposes an algorithm based on deep learning for detecting salient targets in images, which can colorize the video surveillance images of electric power construction and identify the targets in the video surveillance images of electric power construction, but the algorithm has a leakage rate due to the constraints on the recognition range of the algorithm when the targets in the video surveillance images of electric power construction are occluded.

To this end, a new target recognition algorithm based on infrared imaging is proposed. The infrared imaging colorization algorithm based on the local linear mapping method is used to colorize the infrared imaging-based power construction video surveillance image, highlighting the image target, and the target recognition algorithm based on the support vector machine is used to identify the target effectively, and the algorithm has been verified to be effective in the experiments, and can be used as the target recognition problem of power construction video surveillance image. The algorithm is validated in the experiment and can be used as a reference algorithm in the problem of power construction video surveillance image target recognition.

## 2   Target recognition algorithm

### 2.1   Colorization algorithm based on local linear mapping method

The processing flow chart of infrared imaging colorization algorithm based on local linear mapping method is shown in Figure 1.

As shown in Fig. 1, the algorithm first introduces a color reference image, the content and texture of this image are similar to the infrared imaging-based power construction video surveillance image to be processed [j . The local linear mapping method inverts the mapping relationship from the high-dimensional space to the low-dimensional space into the gray-scale space to the color space, and the reconstruction coefficients of the image blocks are calculated in the gray-scale region, based on which the color value of each image block is calculated to transfer the natural color of the template image to the infrared-based power construction video surveillance image to be processed, and then highlight the target through pseudo-color coding [6 -8].

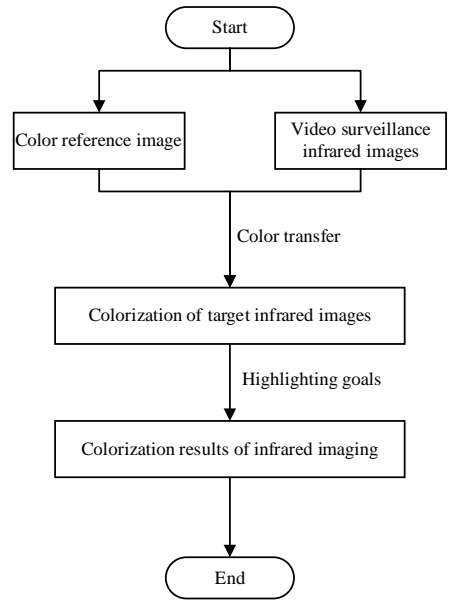

Figure 1. Processing flow chart of infrared imaging colorization algorithm based on local linear mapping method.

### 2.1.1 Color space transformation

The color space transformation is performed by transforming the color template image and the IR-based power construction video surveillance image in the $R$ (red) $G$ (green) $B$ (blue) space to the $YUV$ space, so that the brightness and color information of the IR-based power construction video surveillance image are not connected [9]. The $YUV$ space is a very common chromaticity space for color video information transmission. If only the $Y$ signal component (luminance) exists without the $U$ (chromatic aberration) and $V$ (color difference) components, the image will become a black and white grayscale image. The $U$ (chromatic aberration) and $V$ (chromatic aberration) components are the two components that transform a color image.

The conversion matrix from RGB space to YUV space is:

$$\begin{bmatrix} Y \\ U \\ V \end{bmatrix} = \begin{bmatrix} 0.299 & 0.587 & 0.114 \\ -0.147 & -0.289 & 0.436 \\ 0.615 & -0.515 & -0.100 \end{bmatrix} \begin{bmatrix} R \\ G \\ B \end{bmatrix}$$

**2.1.2 Color transfer**

(1) Preprocessing

Integrate the gray values of each pixel and the pixels around the fixed width $e$ in the IR-based power construction video surveillance image. The vector set $Q_{tar}^j j = 1,2,\cdots,M$, $M$ denotes the number of pixels in the IR-based power construction video surveillance image. Select some pixels in the color template image, and fuse these pixels with the gray values of the surrounding pixels by the same method to obtain a vector set $Q_{tem}^j$.

(2) Nearest Neighbor Search

For each vector in $Q_{tar}^j$, the Euclidean distance calculation method is used to obtain $H$ very close neighbors in $Q_{tem}^j$. The parameter H represents the empirical parameter in the space [10,20].

(3) Operational coefficients

Minimizing each vector in $Q_{tar}^i$ and its $H$ nearest neighbor template vectors and then reconstructing them, the reconstruction result $\alpha(W)$ is

$$\alpha(W) = \sum_{j=1}^{M} \left\| Q_{tar}^j - \sum_{j=1}^{} \bar{\omega}_{ji} Q_{tem}^j \right\|^2$$

where $\bar{\omega}_{ji}$ is the contribution degree.

(4) Color value operation

The color value operation is:

$$d_{tar}^{-j} = \sum_{j=1}^{} \bar{\omega}_{ji} d_{tem}^{-j}$$

where the center pixel color values of $Q_{tar}^j$、 $Q_{tem}^j$ are $d_{tar}^{-j}$、 $d_{tem}^{-j}$, respectively, and the $U$ and $V$ components are included in this color value.

(5) Output result

The obtained color value is fused with the gray value $Y$ of the electric power construction video monitoring image based on infrared imaging, and the colored electric power construction video monitoring image based on infrared imaging is output [10-12].

Convert the image whose output result is $YUV$ space to $RGB$ space, and the transformation matrix from $YUV$ space to $RGB$ space is:

$$\begin{bmatrix} R \\ G \\ B \end{bmatrix} = \begin{bmatrix} 1 & 0 & 1.14 \\ 1 & -0.39 & -0.58 \\ 1 & 2.03 & 0 \end{bmatrix} \begin{bmatrix} Y \\ U \\ V \end{bmatrix}$$

The color transfer of the video surveillance image of power construction based on infrared imaging can be realized by equation (4).

**2.1.3 Pseudo-color coding to highlight the target**

The color difference in color images after color transfer is not as significant as the luminance difference in black and white images, but thermal targets are the main targets in infrared imaging, and the thermal contrast described by the luminance difference will be weakened in the process of color transfer, therefore, pseudo-color coding is used to highlight the thermal targets in the surveillance images after the color transfer [13].

(1) Setting the pixel interval for pseudo-color coding according to the luminance value

According to the color standard, using the $YIQ$ hue coordinates, the image data is divided into a luminance component $Y$, and hue components $I$ and $Q$, which are the image color and saturation, respectively. If the luminance of white light is $Y$, then its red, green and blue light is the relationship between:

$$Y = 0.299R + 0.587G + 0.114B$$

The pseudo-color coding sets the automatic national values according to the difference in the range of luminance values of the thermal targets in the different scenes [14-16]. The pseudo-color coding range is established using the maximum value of the image after color transfer $[t_H, t_{max}]$. $t_H, t_{max}$ are the threshold value and the maximum value of the pixel, respectively.

$$t_H = t_{max} - 0.20(t_{max} - t_H) = 0.80 t_{max} + 0.20 t_{min}$$

where $t_{min}$ is the pixel minimum.

## 2.2 Support Vector Machine Based Target Recognition Algorithm

The support vector machine is used to identify the targets in the colored video surveillance images of electric power construction. A video surveillance image of electric power construction with targets can be regarded as a sample. Set the colored video surveillance image of electric power construction as R, and the features belong to the core of the image. First, the target features are extracted from the image R, which belong to the binary image of the target, and the binary image is subjected to morphological filtering [17]. The details of the target features are shown in Table 1.

The fractal dimension feature $eim$ is a physical quantity that describes the dimensionality of the target space. It can express the density level of strong pixels in a binary map, and is a complementary feature to the standard deviation feature. The fractal dimension only correlates with the density of strong scattering points, but not with the detailed orientation of the scattering points. The feature is:

$$\dot{\text{eim}} = -\frac{\log N_1 - \log N_2}{\log N_1 - \log N_2}$$

where the minimum number of $1 * 1$ pixel windows and the number of $2 * 2$ pixel windows that can cover all bright pixels are, in order, $N_1$、$N_2$.

Set there are $D$ targets in the power construction video surveillance image r to be recognized, and the set of category markers is set to $Z = \{z_1, z_2, \cdots, z_D\}$. The infrared-based power construction video surveillance image $r$ is trained by a support vector machine classifier to obtain a trained recognition classifier [18].

The process of using the classifier to recognize the target of the infrared-based power construction video surveillance image is as follows.

(1) Input the colorized power construction video surveillance image $r$ to implement threshold segmentation, extract the features of the segmented image target, and obtain the feature set of r. Set the infrared-based power construction video surveillance image r there are $D$ kinds of targets to be recognized, $Z = \{z_1, z_2, \cdots, z_D\}$;

(2) Select the feature samples with known kinds of markers in the infrared imaging-based power construction video surveillance image $r$;

Table 1 Target features

| Feature | Meaning | Unit |
|---|---|---|
| Quality | Number of target pixels | - |
| Distance Mean | Mean value of the distance from all target points to the center of the target | m |
| Maximum distance | Maximum value of the distance from all target points to the center of the target | m |
| Minimum value of distance | Minimum value of the distance from all target points to the center of the target | m |
| Diameter | Length of the diagonal with the minimum matrix of the target | m |
| Fractal dimension | Spatial distribution dimension | - |

(3) Operate the hyperplane-based normal vector of the set of feature samples with the optimal hyperplane of deviation $g(r)$:

$$g(r) = z_D r + \text{bias}$$

where the deviation is $bias$.

(4) According to the optimal hyperplane, the target prediction category $C$ can be obtained to realize the target recognition.

$$C = \begin{cases} 1 & g(r) > 0 \\ -1 & g(r) < 0 \end{cases}$$

## 3  Experimental results

According to the actual needs of power construction video surveillance system, we test the timeliness and recognition accuracy of the algorithm for power construction video surveillance image recognition.

The algorithm is used in the video surveillance system of an electric power construction site, and the target in the video surveillance image is set to be a worker, and the surveillance images before and after the algorithm is used in the video surveillance system are shown in Fig. 2 and Fig. 3.

By comparing Figure 2 and Figure 3, it can be seen that before using the algorithm in this paper, the video surveillance image of the power construction site is a black-and-white infrared image, and the target contour is blurred, so it is difficult to identify the target. Before using this algorithm, the video surveillance image of the power construction site is black-and-white infrared image, the target outline is blurred, and the target recognition is difficult; after using this algorithm, the video surveillance image of the power construction site is color infrared image, and the target recognition result is clear and visible. After using this algorithm, the video surveillance image of the power construction site is a colorful infrared image, and the target recognition result is clearly visible.

In order to highlight the recognition effect of this algorithm, the spatial image target recognition algorithm based on non-negative matrix decomposition (literature [3] algorithm), the image based on deep learning significant target detection algorithm (literature [4] algorithm) is also used in the video surveillance system of the electric power construction site, and the recognition effect of the literature [3] algorithm and the literature [4] algorithm is shown in Fig. 4 and Fig. 5.

Fig. 4 and Fig. 5 compare with Fig. 3(b), it can be seen that this algorithm is better at recognizing the targets in the video surveillance image of the electric power construction, and all three staffs are recognized, while the algorithms in the literature [3] and the literature [4] only recognize two staffs, and the algorithm of this algorithm is the most effective in comparison.

The above experiment is to test the indoor power construction video surveillance image target recognition as an example, the next outdoor power construction video surveillance image target recognition as an example, to test this paper's algorithm, the literature [3] algorithm, the literature [4] algorithm of the leakage rate and misrecognition rate, the leakage rate of $\Omega_1$ and the misrecognition rate of $\Omega_2$ is calculated as follows.

$$\Omega_1 = \frac{\gamma_1}{\delta_1} \times 100\%$$

where $\gamma_1$、$\delta_1$ are the number of recognized targets and the number of actual targets, respectively.

$$\Omega_2 = \frac{\gamma_2}{\delta_2} \times 100\%$$

where $\gamma_2$、$\delta_2$ are the number of misrecognized targets and the number of actual targets, respectively.

**Table 2 Training samples and test samples**

| Sample type | Night | | Snowy | | Shelter | | Rainy | |
|---|---|---|---|---|---|---|---|---|
| | training sample (Image)/each | test sample (Image)/each | training sample (Image)/each | test sample (Image)/each | training sample (Image)/each | test sample (Image)/each | training sample (Image)/each | test sample (Image)/each |
| Field personnel | 13 | 11 | 15 | 13 | 16 | 11 | 20 | 18 |
| Power equipment | 11 | 10 | 15 | 12 | 18 | 15 | 20 | 18 |
| Total value | 24 | 21 | 30 | 25 | 34 | 26 | 40 | 36 |

**Table 3 The results of the three algorithms in terms of missed recognition rate (%).**

| Test times | Algorithm in this paper | | | | The algorithm in reference [3] | | | | The algorithm in reference [4] | | | |
|---|---|---|---|---|---|---|---|---|---|---|---|---|
| | Night | Snowy | Shelter | Rainy | Night | Snowy | Shelter | Rainy | Night | Snowy | Shelter | Rainy |
| 1 | 2.01 | 1.99 | 1.76 | 2.01 | 3.43 | 2.43 | 3.43 | 5.34 | 5.43 | 3.24 | 2.09 | 4.32 |
| 2 | 2.12 | 1.98 | 1.76 | 2.11 | 3.44 | 2.66 | 3.33 | 5.55 | 5.98 | 3.43 | 2.34 | 4.21 |
| 3 | 2.09 | 1.76 | 1.76 | 2.09 | 3.65 | 2.32 | 3.24 | 5.35 | 5.66 | 3.42 | 2.32 | 4.23 |
| 4 | 2.08 | 1.76 | 1.76 | 2.06 | 3.48 | 2.76 | 3.46 | 5.34 | 5.86 | 3.43 | 2.35 | 4.23 |
| 5 | 2.13 | 1.89 | 1.75 | 2.04 | 3.45 | 2.31 | 3.43 | 5.55 | 5.64 | 3.45 | 2.65 | 4.24 |
| 6 | 2.32 | 1.67 | 1.64 | 2.05 | 3.45 | 2.87 | 3.46 | 5.35 | 5.66 | 3.42 | 2.35 | 4.11 |

**Table 4 Calculation Results of Misrecognition Rate of Three Algorithms (%)**

| Test times | Algorithm in this paper | | | | The algorithm in reference [3] | | | | The algorithm in reference [4] | | | |
|---|---|---|---|---|---|---|---|---|---|---|---|---|
| | Night | Snowy | Shelter | Rainy | Night | Snowy | Shelter | Rainy | Night | Snowy | Shelter | Rainy |
| 1 | 1.21 | 2.43 | 1.98 | 1.34 | 2.99 | 3.23 | 2.92 | 2.09 | 3.23 | 3.65 | 4.24 | 3.23 |
| 2 | 1.22 | 2.11 | 1.97 | 1.85 | 2.98 | 3.23 | 2.96 | 2.07 | 3.32 | 3.88 | 4.33 | 3.23 |
| 3 | 1.19 | 2.43 | 1.76 | 1.87 | 2.76 | 3.24 | 2.76 | 2.33 | 3.25 | 3.47 | 4.57 | 3.87 |
| 4 | 1.24 | 2.36 | 1.56 | 1.99 | 2.76 | 3.25 | 2.67 | 2.54 | 3.09 | 3.57 | 4.55 | 3.46 |
| 5 | 1.21 | 2.22 | 1.98 | 1.96 | 2.79 | 3.65 | 2.67 | 2.32 | 3.34 | 3.55 | 4.53 | 3.44 |
| 6 | 1.19 | 2.31 | 1.25 | 1.97 | 2.77 | 3.33 | 2.66 | 2.64 | 3.57 | 3.46 | 4.58 | 3.47 |

Four kinds of surveillance environments are set, namely, night, snowy day, sheltered day and rainy day, to test the three algorithms' target recognition effects in the four surveillance environments, and the details of the training samples and test samples during the test are shown in Table 2.

In Table 2, the three algorithms are used to recognize the targets in the monitoring images of electric power construction. The results of the three algorithms are shown in Table 3 and Table 4.

Analyzing Table 3 and Table 4, it can be seen that, among the three algorithms in four monitoring environments, namely, night, snowy day, blockage, and rainy day, this algorithm has the smallest leakage rate and misidentification rate of the video surveillance targets of the electric power construction, and the maximum value of this algorithm's leakage rate and misidentification rate in several tests is 2.32% and the maximum value of this algorithm's misidentification rate is 2.43%. Compared with the algorithms in literature [3] and [4], the recognition performance of this algorithm is the best in the four surveillance environments of night, snow, cover and rain.

## 4 Conclusion

In-depth study of a power construction video surveillance image target recognition algorithm based on infrared imaging, the algorithm through the monitoring image preprocessing, monitoring image target recognition two steps to realize the power construction video surveillance image target recognition, in the experiment, the algorithm has been verified as follows.

(1) The algorithm has a good effect on the power construction video surveillance image target recognition, all three staff members are recognized, and the recognition effect is better than the comparison algorithm;

(2) In the four monitoring environments of night, snow, shelter and rain, this algorithm minimizes the leakage rate and misidentification rate of the power construction video surveillance target, and the leakage rate and misidentification rate are lower than that of the comparison algorithm, so the recognition effect is the best.

**References:**

[1] Wang, L., & Liu, Q. (2018). Multi-objective Image Segmentation Algorithm Based on Local Feature H2 Characteristic. Progress in Laser and Optoelectronics, 55(06), 109-116.
[2] Li, X., Shi, Y., Song, Y., et al. (2017). Research on Near Infrared Imaging System Based on AOTF. Piezoelectrics & Acoustooptics, 39(02), 289-294.
[3] Sun, J., & Zhao, F. (2019). Application of Non-negative Matrix Factorization in Spatial Object Image Recognition. Progress in Laser and Optoelectronics, 56(10), 122-129.
[4] Zhao, H., & An, W. (2018). Image Salient Object Detection Combining Deep Learning. Progress in Laser and Optoelectronics, 55(12), 192-200.
[5] Chen, Q., Hu, H., & Liu, L. (2019). Study on Collection and Feature of Laser-Arc Hybrid Welding Pool Images. Applied Laser, 39(05), 825-834.
[6] Zhou, L., & Liu, Y. (2019). Laser Image Processing Based on Adaptive Genetic Algorithm. Journal of Shenyang University of Technology, 41(02), 174-178.
[7] Qin, Y., Wu, J., & An, W. (2017). Feature Extraction of Laser Grid Marker Image Based on RANSAC. Computer Engineering and Science, 39(08), 1495-1501.
[8] Song, R., Zhang, H., Meng, F., et al. (2017). Research on Denoising Algorithm of Laser Active Imaging Guidance Image. Optoelectronics & Control, 24(10), 7-11+26.
[9] Wang, M., Fan, Y., Chen, B., et al. (2017). Design of Infrared Image Adaptive Non-uniformity Correction Based on SOPC. Infrared and Laser Engineering, 46(06), 208-213.
[10] Lv, S., Yang, F., Ji, G., et al. (2018). Multiclass Multimodal Combination Fusion of Infrared Intensity and Polarization Images. Infrared and Laser Engineering, 47(05), 63-72.
[11] Wang, C., Tang, X., & Gao, S. (2017). Research on Infrared Image Enhancement Algorithm Based on Human Visual Perception. Laser & Infrared, 47(01), 114-118.
[12] Zhu, H., Liu, C., Zhang, B., et al. (2018). Research on Laser Ultrasonic Visualization Image Processing. China Laser, 045(01), 168-175.
[13] Gu, G., & Liu, L. (2017). Infrared Image Enhancement Based on Improved Top-Hat Transform of Rail Crack. Laser & Infrared, 47(01), 47-52.
[14] Wei, L., Ding, M., Zeng, L., et al. (2017). Pedestrian Detection in Infrared Images under Laser Radar Guidance. Computer Engineering and Application, 053(023), 197-202.
[15] Nie, F., Li, J., Zhang, P., et al. (2017). Kaniadakis Entropy Threshold Segmentation Method for Complex Images. Laser & Infrared, 47(08), 1040-1045.
[16] Feng, X., Jiang, C., He, M., et al. (2019). Adaptive Smoothing of Three-dimensional Distance Images Based on Feature Estimation. Optics and Precision Engineering, 27(12), 2693-2701.
[17] Zhao, Y., & Xu, K. (2017). Three-dimensional Reconstruction Algorithm of Laser Image Feature Based on Constrained Fluidization. Machine Tool & Hydraulic, 45(018), 34-40.
[18] Liu, W., Zhang, Y., Gao, P., et al. (2017). Subpixel Center Extraction Method of Laser Stripe Based on Hierarchical Processing.

Infrared and Laser Engineering, 46(10), 232-239.

[18] Zhou, Qiangqiang, and Zhenbing Zhao. "Substation equipment image recognition based on SIFT feature matching." 2012 5th International Congress on Image and Signal Processing. IEEE, 2012.

[19] Liu, Yunpeng, et al. "Research on automatic location and recognition of insulators in substation based on YOLOv3." High Voltage 5.1 (2020): 62-68.

[20] Jiang, Anfeng, et al. "Visible image recognition of power transformer equipment based on mask R-CNN." 2019 IEEE Sustainable Power and Energy Conference (iSPEC). IEEE, 2019.

[21] Xiong, Siheng, et al. "Object recognition for power equipment via human-level concept learning." IET Generation, Transmission & Distribution 15.10 (2021): 1578-1587.

[22] Liu, Yang, Jun Liu, and Yichen Ke. "A detection and recognition system of pointer meters in substations based on computer vision." Measurement 152 (2020): 107333.

[23] Jiang, Anfeng, et al. "Research on infrared image recognition method of power equipment based on deep learning." 2020 IEEE International Conference on High Voltage Engineering and Application (ICHVE). IEEE, 2020.

[24] Xu, Liang, et al. "An efficient foreign objects detection network for power substation." Image and Vision Computing 109 (2021): 104159.